\documentclass{sig-alternate-MLSDA14}
\usepackage{multirow}
\usepackage{amsmath,bm}
\usepackage{graphicx,dblfloatfix} 
\usepackage{balance}
\usepackage[pdfauthor={Johanna Carvajal, Conrad Sanderson, Chris McCool, Brian C. Lovell},pdftitle={Multi-Action Recognition via Stochastic Modelling of Optical Flow and Gradients},pagebackref=true,breaklinks=true,letterpaper=true,colorlinks,citecolor=blue,urlcolor=blue,linkcolor=blue,bookmarks=false]{hyperref}

\graphicspath{{.}}

\begin{document}

\toappear
  {
  \hrule\vspace{2ex}
  Published in:  Workshop on Machine Learning for Sensory Data Analysis (MLSDA), pp.~19-24, 2014.\\
  \href{http://dx.doi.org/10.1145/2689746.2689748}{http://dx.doi.org/10.1145/2689746.2689748}
  }

\title{Multi-Action Recognition via Stochastic Modelling of Optical Flow and Gradients}
\numberofauthors{1}
\author{
Johanna Carvajal, Conrad Sanderson, Chris McCool, Brian C. Lovell\\
~\\
~\\
\affaddr{NICTA, 70 Bowen St, Spring Hill, QLD 4000, Australia}\\
\affaddr{University of Queensland, St Lucia, QLD 4072, Australia}\\
\affaddr{Queensland University of Technology (QUT), Brisbane, QLD 4000, Australia}\\
~\\
}

\maketitle

\begin{abstract}

\vspace{1ex}

In this paper we propose a novel approach to multi-action recognition that performs joint segmentation and classification.
This approach models each action using a Gaussian mixture using robust low-dimensional action features.
Segmentation is achieved by performing classification on overlapping temporal windows, which are then merged to produce the final result.
This approach is considerably less complicated than previous methods which use dynamic programming or computationally expensive hidden Markov models (HMMs).
Initial experiments on a stitched version of the KTH dataset show that the proposed approach achieves an accuracy of 78.3\%,
outperforming a recent HMM-based approach which obtained 71.2\%.

\end{abstract}

\terms{Algorithms, Performance, Experimentation}

\vspace{2ex}

\section{Introduction}

\vspace{1ex}

Action recognition in real world surveillance videos has become of great interest due to its potential applications in daily life  situations such as smart homes, home nursing, building security, and annotation of human actions in video using minimal manual supervision~\cite{Duchenne2009}.

Human action recognition can be divided into two areas: {\bf (i)} single action recognition, and {\bf (ii)} multi-action recognition. In the context of this paper we define single action recognition as the unique action performed by a person, and multi-action recognition as a set of  actions where one person performs a \textit{sequence} of such single actions~\cite{Qinfeng2008}.

In most computer vision literature, action recognition approaches have concentrated on recognising a single action performed by a person, instead of continuous human actions or multi-actions~\cite{Buchsbaum2011}. 
Multi-action recognition is a fundamental problem in human action understanding. When observing videos of human behaviour, an ongoing problem for computer vision algorithms is recognising and/or segmenting individual and significant actions from within the duration of the motion sequence~\cite{Buchsbaum2011}. It is challenging due to the high variability of appearances, shapes, possible occlusions, large variability in the temporal scale and periodicity of human actions, the complexity of articulated motion, the exponential nature of all possible movement combinations, as well as the prevalence of irrelevant background~\cite{Hoai2011,Qinfeng2008}.

Limited work has been conducted on multi-action recognition. A joint action segmentation and classification method was presented in~\cite{Shimosaka2007} called Multi-Task Conditional Random Field (MT-CRF) which classifies motions into multiple labels, such as a person folding their arms while seated. However, this approach has only been applied to synthetic datasets. Hoai et al.~\cite{Hoai2011} address joint segmentation and classification by classifying temporal regions using a multi-class SVM and performing segmentation using dynamic programming. More recently, Borzehsi et al.~\cite{Borzeshi2013} proposed the use of hidden Markov models with irregular observations (termed HMM-MIO) to perform multi-action recognition. A drawback for both~\cite{Borzeshi2013} and~\cite{Hoai2011} is that they have a large number of parameters to optimise. Furthermore, \cite{Borzeshi2013} uses very high dimensional feature vectors and~\cite{Hoai2011} requires fully labelled annotations for training.

\begin{figure}[!b]
  \centering
    \includegraphics[width=0.4\textwidth]{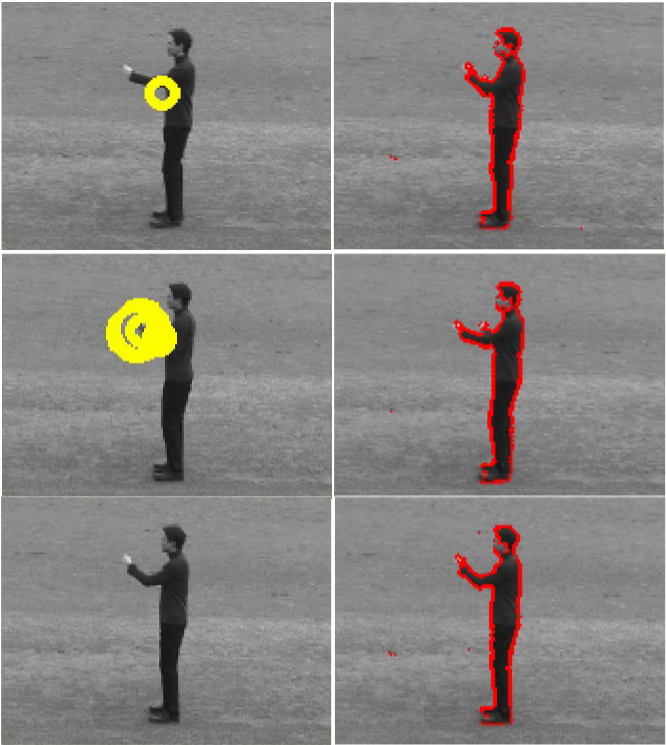}
    \caption{Left: Spatio-Temporal Interest Points Descriptors (STIPs) are unstable, imprecise and overly sparse. Right: interest pixels (marked in red) obtained using magnitude of gradient.}
\label{fig:Descriptors}
\end{figure}

A reliable feature descriptor is a crucial stage for the success of an action  recognition system. One popular descriptor for the action recognition task is Spatio-Temporal Interest Points (STIPs)~\cite{Laptev2005}.
STIP based descriptors have some drawbacks~\cite{Liangliang2010,Chakraborty2012,Kliper2012, ChuanZhen2014}:
(i)~they focus on local spatio-temporal information instead of global motion,
(ii)~they can be unstable and imprecise (varying number of STIP detections) leading to low repeatability,
(iii)~they are computationally expensive,
and
(iv)~produce sparse detections.
See Fig.~\ref{fig:Descriptors} for a demonstration of STIP based detection.

Other feature extraction techniques used for action recognition include gradients~\cite{KReddy2013} and optical flow~\cite{Ali2010,Kliper2012}.
Each pixel in the gradient image helps extract relevant information, e.g.~edges. Since the task of action recognition is based on a sequence of frames,
optical-flow provides an efficient way of capturing the local dynamics in the scene~\cite{Kliper2012}.
See the right side in Fig.~\ref{fig:Descriptors} for a demonstration of gradient based detection.

In this paper we propose a novel framework to perform multi-action recognition that requires few parameters. We model each action as a Gaussian mixture model (GMM) using the action features described in~\cite{AndresSanin2013} and use as training data the videos of single actions. We use robust low-dimensional action features which incorporate optical flow and image gradient information and we only use those features which have high spatial frequency (that correspond to edges) to be considered part of an action. 
In this way we only use relevant action information. Segmentation of actions is achieved by applying the GMM classifier over a temporal sliding window in an overlapping manner, which allows us to better deal with temporal misalignment and can lead to improved performance~\cite{KaiGuo2013}.

\textbf{Contributions.}  We propose a more efficient system that requires fewer parameters to be optimised, and simultaneously segments and recognises multi-actions.
This is in contrast to the methods presented in~\cite{Borzeshi2013, Hoai2011} which have a larger parameter search space.
Furthermore, we avoid the need for a custom dynamic programming definition. Lastly, the feature descriptors used in this work are more robust and have smaller dimensionality. The proposed method is able to handle varying number of feature vectors obtained from each frame, allowing selective feature extraction from the most useful image areas.

We continue the paper as follows. In Section~\ref{sec:related}, we summarise prior work for feature descriptors, single action and multi-action recognition. We then describe the proposed method for joint multi-action segmentation and recognition in Section~\ref{sec:proposed_method}. In Section~\ref{sec:experiments}, we present experiments which show that the proposed method outperforms existing methods. Section~\ref{sec:conclusions} summarises the main findings and provides potential areas for future work.

\newpage
\section{Related Work}\label{sec:related}

In this section, we present an overview of the state-of-the-art of action recognition. To begin with, we depict some popular descriptors used for action recognition. This is followed by a review of techniques for single action recognition. Finally,  we present an overview of the approaches which perform multi-action segmentation and recognition.

\subsection{Descriptors for Action Recognition} \label{subsec:descriptors}

Several descriptors have been used to represent human actions. Among them, we can find descriptors based on Spatio-Temporal Interest Points (STIPs)~\cite{Laptev2005}, gradients~\cite{KReddy2013}, and optical flow~\cite{Ali2010}.
The STIP based detector finds interest points and is a space-time extension of the Harris and F\"{o}rstner interest point detector~\cite{Fostner87,Harris88}. Recently, STIP descriptors have led to relatively good action recognition performance~\cite{Borzeshi2012}. However, STIP based descriptors have some drawbacks as reported in~\cite{Liangliang2010,Chakraborty2012,Kliper2012, ChuanZhen2014}.  These drawbacks include: focus on local spatio-temporal information instead of global motion, can be unstable and imprecise (varying number of STIP detections) leading to low repeatability, redundancy can occur, are computationally expensive and the detections can be overly sparse (see Fig.~\ref{fig:Descriptors}).
  
Gradients have been used as a robust image and video representation~\cite{KReddy2013}. Each pixel in the gradient image helps extract relevant information, e.g. edges. Gradients can be computed at every spatio-temporal location $(x, y, t)$ in any direction in a video.

Since the task of action recognition is based on a sequence of frames in order to analyse the various motion patterns in the video~\cite{KReddy2013}, optical-flow provides an efficient way of capturing the local dynamics in the scene~\cite{Kliper2012}. Optical flow describes the motion dynamics of an action, calculating the absolute motion between two frames, which contains motion from many sources~\cite{Ali2010,KaiGuo2013,HengWang2013Journal}. 

\subsection{Single Action Recognition}

Hidden Markov models (HMMs) have a long history of use in activity recognition. One action is a sequence of events ordered in space and time, and HMMs capture structural and transitional features and therefore the dynamics of the system~\cite{Mendoza2007}. 
Gaussian Mixture Models (GMMs) have also been explored for recognising single actions. In~\cite{Liangliang2010}, each set of feature vectors is modelled with a GMM. Then, the likelihood of a feature vector belonging to a given human action can be estimated.

Other successful methods for single action recognition include Riemannian manifold based approaches and those that use dense trajectory tracking.
For single action recognition, Riemannian manifolds have been investigated in~\cite{KaiGuo2010,AndresSanin2013}. In~\cite{KaiGuo2010} optical flow features are extracted and then a compact covariance matrix representation of such features is calculated. Such a covariance matrix can be thought of as a point on a Riemannian manifold.
An action and gesture recognition method based on spatio-temporal covariance descriptors obtained from optical flow and gradient descriptors is presented in~\cite{AndresSanin2013}. 

Two recent approaches for single action recognition based on tracking of interest points makes use of dense trajectories~\cite{HengWang2013Journal,HengWang2013iccv}. These dense trajectories allow the description of videos by sampling dense points from each frame and then tracking them based on displacement information from a dense optical flow field. Although this approach obtains good performance, it is computationally expensive, especially the calculation of the dense optical flow which is calculated at several scales. 

\subsection{Multi-Action Recognition}

Multi-action recognition, in our context, consists of segmenting and recognising single actions from a video sequence where one person performs a sequence of such actions~\cite{Qinfeng2008}.
The process for segmenting and recognising multiple actions in a video can be solved either as two independent problems or a joint problem.

One of the first methods to multi-action recognition,  Multi-Task Conditional Random Field (MT-CRF), was proposed in~\cite{Shimosaka2007}. This method consists of classifying motions into multi-labels, e.g.~a person folding their arms while sitting. Despite this approach being presented as robust, it has been only applied on two synthetic datasets.
Two more recent methods~\cite{Hoai2011,Borzeshi2013} have been applied to more realistic multi-action datasets.
Hoai et al.~\cite{Hoai2011} deal with the dual problem of human action segmentation and classification. This approach  is presented as a learning framework that simultaneously performs temporal segmentation and event recognition in time series.
The supervised training is done via multi-class SVM, where the SVM weight vectors have to be learnt, as well as the other SVM parameters. For the segmentation, the learnt weight vectors are used and other set of parameters are optimised (number of segments, and minimum and maximum lengths of segments). The segmentation is done using dynamic programming. The feature mapping depends on the dataset employed, and includes trajectories, features extracted from binary masks and STIPs.

Although the method proposed in~\cite{Hoai2011} is promising, it has several drawbacks. One drawbacks is the requirement of fully labelled annotation for training.  Furthermore, it suffers from the limitations of dynamic programming where writing the code that evaluates sub-problems in the most efficient order is often nontrivial~\cite{Wagner1995}. Also, the binary masks are not always available and the STIPs descriptors present deficiencies as mentioned in Section~\ref{subsec:descriptors}. This method also requires an extensive search for optimal parameters.

An approach termed Hidden Markov Model for Multiple, Irregular Observations (HMM-MIO)~\cite{Borzeshi2013} has also been proposed for the multi-action recognition task.
HMM-MIO jointly segments and classifies observations which are irregular in time and space, and characterised by high dimensionality.  The high dimensionality is reduced by using PPCA (probabilistic Principal Component Analysis). Moreover, HMM-MIO deals with heavy tails and outliers exhibited by empirical distributions by modelling the observation densities with a long-tailed distribution, the student's $t$. HMM-MIO requires the search of the following 4 optimal parameters: (i) the resulting reduced dimension ($R$), (ii) the number of components in each observation mixture ($M$), (iii) the degree of the $t$-distribution ($v$), and (iv) the number of cells (or regions) used to deal with the space irregularity $S$. To this end, for multiple observations of a frame, they postulate:

\noindent
\begin{equation}
p(x_t^{1:N_t}|y_t) = 
\left\{\begin{matrix}
 \prod_{n=1}^{N_t} p(x_t^n|y_t) & \text{if} \,N_t \geq  1 \\  \\
 1 & \text{if} \,N_t = 0  
\end{matrix}\right.
\end{equation}

\noindent
where each observation consists of the pair $x_t^n=\{d_t^n,s_t^n\}$ of the descriptor $d_t^n$ and the cell index where it occurs $s_t^n$. The index frame, the observation index, the total number of observations, and the hidden states are given by $t$, $n$, $N_t$, and $y_t$, respectively. As feature descriptors, HMM-MIO extracts STIPs proposed in~\cite{Laptev2005}, with the default $162$-dimension descriptor. The classification is carried out on a per frame basis.  HMM-MIO also suffers from the drawback of a large search of optimal parameters and the use of STIP descriptors.

\section{Proposed Method}
\label{sec:proposed_method}

\subsection{Selective Feature Extraction}

Descriptors based on optical flow and gradient have been proved to be reliable for the action recognition task~\cite{KaiGuo2013,AndresSanin2013}.
Following~\cite{AndresSanin2013}, we extract the following features for each pixel:
\begin{equation}\label{eq:features}
\bm{f}(x,y,t)  := [\; x, \; y, \; \bm{g}, \; \bm{o} \;]^T
\end{equation}

\noindent
where $x$ and $y$ are the pixel coordinates, while $\bm{g}$ and $\bm{o}$ are defined as:
\begin{eqnarray}
\hspace{-2ex} \bm{g} \hspace{-1ex} & = & \hspace{-1ex}
\left[ \; |J_x|, \; |J_y|, \; |J_{yy}|, \; |J_{xx}|, \; \sqrt{J_x^2 + J_y^2}, \; \text{atan} \frac{|J_y|}{|J_x|} \; \right]
\label{eq:features2}\\
\hspace{-2ex} \bm{o} \hspace{-1ex} & = & \hspace{-1ex}
\left[\; u, \;\;  v, \;\; \frac{\partial{u}}{\partial{t}}, \;\;  
\frac{\partial{v}}{\partial{t}}, \;\; 
\left (\frac{\partial u}{\partial x} + \frac{\partial v}{\partial y} \right ), \;\;
\left (\frac{\partial v}{\partial x} - \frac{\partial u}{\partial y} \right ) \;
\right]
\label{eq:features3}
\end{eqnarray}

The first four gradient-based features in Eq.~(\ref{eq:features2}) represent the first and second order intensity gradients at pixel location $(x,y)$. The last two gradient features represent gradient magnitude and gradient orientation. The optical-flow based features in Eq.~(\ref{eq:features3}) represent in order: the horizontal and vertical components of the flow vector, the first order derivatives with respect to $t$, the divergence and vorticity of the optical flow in the context of action recognition proposed in~\cite{Ali2010}. We obtain a 14-dimensional feature vector per pixel.

Due to not all pixels corresponding to the object of interest, we are only interested in pixels with a gradient magnitude greater than a threshold~$\tau$~\cite{KaiGuo2013}.
As such, we discard all the features vectors from locations with a small magnitude.
See the right side of Fig.~\ref{fig:Descriptors} for an example.

\vspace{1ex}
\subsection{Learning Action Models}
\label{subsec:train}

A Gaussian Mixture Model (GMM) is a weighted sum of $N_g$ component Gaussian densities~\cite{Bishop_PRML_2006}, defined as:
\begin{equation}
p(\bm{x} | \lambda) = \sum\nolimits_{g=1}^{N_g}w_g\mathcal{N}(\bm{x}|,\bm{\mu}_g, \bm{\Sigma}_g)
\end{equation}

\noindent
where $\bm{x}$ is a $D$-dimensional continuous-valued data vector (i.e. measurement or features), $w_g$ is the weight of the $g$-th Gaussian (with constrains $0\leq w_g \leq 1$ and $\sum_{g=1}^{N_g}w_g=1$), and $\mathcal{N}(\bm{x}|,\bm{\mu}_g, \bm{\Sigma}_g)$ is the component Gaussian density with mean $\bm{\mu}$ and covariance matrix $\bm{\Sigma}$, given by: 
\begin{equation*}
\mathcal{N}(\bm{x}|,\bm{\mu}, \bm{\Sigma}) \mbox{=} \frac{1}{(2\pi)^{\frac{D}{2}}|\bm{\Sigma}|^{\frac{1}{2}}}
\exp {\left\{ -\frac{1}{2}(\bm{x}-\bm{\mu})^T\bm{\Sigma}^{-1}(\bm{x}-\bm{\mu}) \right\}}
\end{equation*}

The complete Gaussian mixture model is parametrised by the mean vectors, covariance matrices and weights
of all component densities. These parameters are collectively represented by the notation
$\lambda= \{w_g,\bm{\mu}_g,\bm{\Sigma}_g\}_{g=1}^{N_g}$.

Each video $V_i$ is represented as a set of frames $\{I_t\}_{t=1}^T$ and a set of corresponding action labels $\{l_t\}_{t=1}^T$.
Each frame is represented as a set of $k$ feature vectors \mbox{$F_t=\{\bm{f}_t^1,\bm{f}_t^2,\cdots,\bm{f}_t^k\}$},
where $k$ can vary depending on frame content.

All the feature vectors belonging to the same action are pooled together.
For each action a GMM is trained with $N_g$ components. This results in a set of GMM models that we will express as $\{ \lambda_a \}_{a=1}^A $, where $A$ is the total number of actions.

\subsection{Recognition of Multiple Actions}

For each frame $I_t$ in a given testing video $V_i$, we calculate the feature vectors $\bm{f}(x,y,t)$ following the same procedure as explained in Section~\ref{subsec:train}. Each frame is then represented as a set of $k$ feature vectors $F_t=\{\bm{f}_t^1,\bm{f}_t^2,\cdots,\bm{f}_t^k\}$. We break $V_i$ into small overlapping segments and classify each of these segments. 

As reported in~\cite{KaiGuo2013}, many human actions are repetitive in nature, such as walking and boxing. An important consideration is to determine the duration of each segment $L$ such that it is long enough to contain one complete cycle of the action. 

Let $X^s = [ F_t \, F_{t+1} \cdots F_{t+L}  ]$ be the sequence of $N$ feature vectors taken from segment $s_{(t,t+L]}$. The feature vectors $X^{s}$ are assumed independent, so the average log-likelihood of a model $\lambda_a$ is computed as:
\begin{equation}\label{eq:log_like}
\log p(X^s|\lambda_a) = \frac{1}{N}\sum\nolimits_{i=1}^N  \log p(\bm{f}_i|\lambda_a).
\end{equation}

In each segment $s$, we compute the average log-likelihood for each model $\lambda_a$ by using Eq.~(\ref{eq:log_like}). A log-likelihood vector $\widehat{\textbf{p}}_s$ is obtained:
\begin{equation}
\widehat{\textbf{p}}_s = \left[ ~ \log p(X^s|\lambda_1), ~ \cdots, ~ \log p(X^s|\lambda_A) ~ \right]. 
\end{equation}

Then,  a new vector of log-likelihoods $\widehat{\textbf{p}}_{s+1}$ is calculated in the next segment, jumping one frame at a time. We repeat this process until we reach the end of the video, resulting in a set of $S$ log-likelihood vectors $\{\widehat{\textbf{p}}_s\}_{s=1}^S$.

To calculate the total contribution $\textbf{p}_\text{sum}^{F_t} $ of the log-likelihood vectors to each frame, we sum all the $\widehat{\textbf{p}}_s$ obtained over this frame: 
\begin{equation}
\textbf{p}_\text{sum}^{F_t} = \sum\nolimits_{\widehat{\textbf{p}}_s \in F_t} \widehat{\textbf{p}}_s 
\end{equation}

\noindent
The estimated label $\widehat{l}_t$ is calculated as:
\begin{equation}
\widehat{l}_t = \underset{\forall \lambda_a} {\mathrm{argmax}} ~\textbf{p}_\text{sum}^{F_t}
\end{equation}

We aim to segment the video into separate actions. To this end, we examine the sequence of estimated labels. Knowing that the duration of each segment is $L$, if any estimated sequence has a length less than $L$, the short segment is considered a transition boundary between two actions and for this reason this estimated sequence is combined to the preceding segment.

\section{Experiments}\label{sec:experiments}

\begin{figure}[tb!]
  \centering
    \includegraphics[width=0.24\columnwidth]{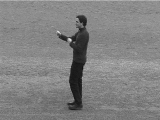}
    \includegraphics[width=0.24\columnwidth]{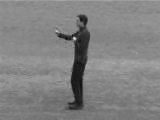}
    \includegraphics[width=0.24\columnwidth]{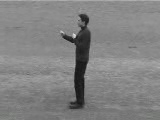}
    \includegraphics[width=0.24\columnwidth]{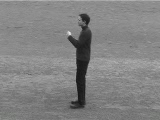}
    
    \vspace{-2ex}
    \caption{The ``boxing'' action in the KTH dataset.}
    \label{fig:boxing_action}
\end{figure}

We evaluated our proposed method for joint action segmentation and classification on a stitched version of the KTH dataset~\cite{Schuldt2004}. The KTH dataset is composed of 6~types of human actions (walking, jogging, running, boxing, hand waving and hand clapping) which are performed several times by 25 subjects in 4 scenarios: outdoors, outdoors with scale variation, indoors and outdoors with varying clothes. The image size is of $160 \times 120$
pixels, and temporal resolution is of 25 frames per second. In total there are $25\times 6 \times 4 =600$ video files. Each video contains an individual performing the same action. See Fig.~\ref{fig:boxing_action} for an example. This action is performed 4 times and each subdivision or action-instance (in terms of \textit{start-frame} and \textit{end-frame}) is provided as part of the dataset\footnote{{http://www.nada.kth.se/cvap/actions/}}. This dataset contains $2391$ action-instances, with a length between $1$ and $14$ seconds~\cite{Baccouche2011}.

\begin{figure}[tb!]
  \centering
    \includegraphics[width=0.24\columnwidth]{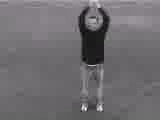}
    \includegraphics[width=0.24\columnwidth]{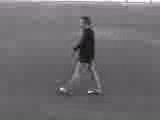}
    \includegraphics[width=0.24\columnwidth]{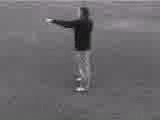}
    \includegraphics[width=0.24\columnwidth]{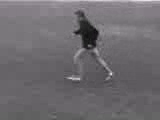}
    
    \vspace{-2ex}
    \caption{Multi-action sequence in the stitched version of the KTH dataset: hand clapping, walking, boxing and running.}
    \label{fig:stiched_example}
\end{figure}

Following~\cite{Borzeshi2013}, the stitched dataset was obtained by simply concatenating existing single-action instances into sequences. The action-instances were picked randomly alternating between the two groups of \{boxing, hand-waving, and hand-clapping\} and \{walking, jogging, and running\} to accentuate action boundaries. We refer to each of these sequences a ``multi-action video''.
See Fig.~\ref{fig:stiched_example} for an example.

The dataset was divided into two sets: one for training and one for testing. $64$~and $36$ multi-action videos were used for training and testing respectively. For the experiment, we used $3$-fold cross-validation. We use the original dataset to train our GMM model, it means that we only use video containing single actions to train each GMM per action. As the KTH dataset was collected in $4$ scenarios, we train a model per action and per scenario. Each pixel descriptor is a $14$-dimensional feature vector and it is extracted using Eq.~(\ref{eq:features}). All the feature vectors belonging to a frame $t$ is given by $F_t$. Although the optical flow is calculated in all frames, in order to speed up processing, we only use the feature vectors extracted from second frame.

For our experiments, we used diagonal covariance matrices. GMM parameters were estimated using descriptors obtained from training videos using the iterative Expectation-Maximisation (EM) algorithm~\cite{Bishop_PRML_2006}.
The experiments were implemented with the help of the Armadillo C++ library~\cite{Sanderson_NICTA_2010}.
The threshold $\tau$ (used for selecting feature vectors) was empirically set to $40$. The duration of each segment $L$ was set to $25$ frames ($1$~second), which is the minimum length of an action-instance in this dataset~\cite{Baccouche2011}.

An initial set of experiments was performed to find the optimal number of components $N_g$. Using a fixed number of components $N_g=\{ 16, 64, 256, 1024 \}$, we evaluate the performance on one fold.
The results are reported in frame-level accuracy ($\%$) in Table \ref{table:com_Ng}.

We found that using $N_g=1024$  provided optimal performance ($77.0\%$). This better performance attained with $1024$ components is explained by the fact that  GMMs with large number of components are known to have the ability to model any given probability distribution function~\cite{Liangliang2010}. 
We kept the number of Gaussians constant for the remainder of our experiments.

\begin{table}[!t]
\centering
\caption{Comparison of one run testing for several number of Gaussians ($N_g$).}
\vspace{1ex}
\begin{tabular}{|c|c|c|c|c| }
\hline
{~~~~$N_g$~~~~} & {\bf Accuracy (\%)}  \\ \hline
  16 & 71.1  \\ \hline
  64 & 73.2  \\ \hline
 256 & 75.3  \\ \hline
1024 & 77.0  \\ \hline
\end{tabular}
\label{table:com_Ng}
\end{table}

We compared our proposed method with the state-of-the-art HMM-MIO method~\cite{Borzeshi2013}. As we mentioned in Section~\ref{sec:related}, HMM-MIO requires the search of many optimal parameters. This results in a complex space parameter search ($4$ free parameters). In contrast, our proposed method only has $2$ free parameters ($\tau$ and $Ng$).

The comparative results of the proposed method, the HMM-MIO, and the baseline system Bag-of-Features approach trained with k-means using $256$ clusters, are shown in Table~\ref{table:results}. The proposed method obtains the highest accuracy ($78.3 \pm 2.6 \%$). We report the results for proposed method in terms of average accuracy and standard deviation over the 3 folds.

\begin{table}[!t]
\centering
\caption{Comparison of the proposed method to the state of the art.}
\vspace{1ex}
\begin{tabular}{|c|c| }
\hline
\textbf{Method} & \textbf{Accuracy} (\%)\\ \hline
Bag-of-features~\cite{Borzeshi2013} & $61.8$ \\ \hline
HMM-MIO~\cite{Borzeshi2013} & $71.2$ \\ \hline
proposed method &  $78.3 \pm 2.6 $\\ \hline  
\end{tabular}
\label{table:results}
\end{table}
\section{Conclusions and Future Work}\label{sec:conclusions}

In this paper, we have proposed an improved approach for joint multi-action segmentation and recognition from video sequences. We proposed the use of low-dimensional gradient and optical flow based descriptors which do not suffer from the instability, imprecision and sparsity exhibited by STIP descriptors used by the HMM-MIO system of~\cite{Borzeshi2013}. The proposed approach also obviates the need for an extra dimensionality reduction step, as is the case for the HMM-MIO system. Furthermore, the proposed system provides a simpler framework with half the number of parameters to optimise.
Initial experiments on a stitched version of the KTH dataset show that the proposed approach achieves an accuracy of 78.3\%,
compared to 71.2\% achieved by the HMM-MIO system.

Possible directions for future work are: {\bf (i)} explore the use of spatio-temporal descriptors dividing each frame into cells or regions to deal with spatial irregularities,
and {\bf (ii)}~separate irrelevant motion from action motion via explicit foreground segmentation~\cite{Reddy_TCSVT_2013}, which would be especially useful when dealing with actions in uncontrolled settings.

\section{Acknowledgments}

NICTA is funded by the Australian Government through the Department of Communications,
as well as the Australian Research Council through the ICT Centre of Excellence Program.

\small
\newpage
\balance
\bibliographystyle{abbrv}

\bibliography{references} 

\begin{thebibliography}{10}

\bibitem{Ali2010}
S.~Ali and M.~Shah.
\newblock Human action recognition in videos using kinematic features and
  multiple instance learning.
\newblock {\em Pattern Analysis and Machine Intelligence}, 32(2):288--303,
  2010.

\bibitem{Baccouche2011}
M.~Baccouche, F.~Mamalet, C.~Wolf, C.~Garcia, and A.~Baskurt.
\newblock Sequential deep learning for human action recognition.
\newblock In {\em Human Behavior Understanding}, volume 7065 of {\em Lecture
  Notes in Computer Science}, pages 29--39. Springer Berlin Heidelberg, 2011.

\bibitem{Bishop_PRML_2006}
C.~M. Bishop.
\newblock {\em Pattern Recognition and Machine Learning}.
\newblock Springer, 2006.

\bibitem{Borzeshi2013}
E.~Borzeshi, O.~Perez~Concha, R.~Xu, and M.~Piccardi.
\newblock Joint action segmentation and classification by an extended hidden
  {M}arkov model.
\newblock {\em IEEE Signal Processing Letters}, 20(12):1207--1210, 2013.

\bibitem{Buchsbaum2011}
D.~Buchsbaum, K.~R. Canini, and T.~Griffiths.
\newblock Segmenting and recognizing human action using low-level video
  features.
\newblock {\em Annual Conference of the Cognitive Science Society}, 2011.

\bibitem{Liangliang2010}
L.~Cao, Y.~Tian, Z.~Liu, B.~Yao, Z.~Zhang, and T.~Huang.
\newblock Action detection using multiple spatial-temporal interest point
  features.
\newblock In {\em International Conference on Multimedia and Expo (ICME)},
  pages 340--345, 2010.

\bibitem{Chakraborty2012}
B.~Chakraborty, M.~B. Holte, T.~B. Moeslund, and J.~Gonz\'{a}lez.
\newblock Selective spatio-temporal interest points.
\newblock {\em Computer Vision and Image Understanding}, 116(3):396--410, 2012.

\bibitem{Duchenne2009}
O.~Duchenne, I.~Laptev, J.~Sivic, F.~Bach, and J.~Ponce.
\newblock Automatic annotation of human actions in video.
\newblock In {\em International Conference on Computer Vision (ICCV)}, pages
  1491--1498, 2009.

\bibitem{Fostner87}
M.~A. F\"{o}stner and E.~G\"{u}lch.
\newblock A fast operator for detection and precise location of distinct
  points, corners and centers of circular features.
\newblock In {\em ISPRS Intercommission Workshop}, 1987.

\bibitem{Kliper2012}
O.~K. Gross, Y.~Gurovich, T.~Hassner, and L.~Wolf.
\newblock Motion interchange patterns for action recognition in unconstrained
  videos.
\newblock In {\em European Conference on Computer Vision ({ECCV})}, 2012.

\bibitem{KaiGuo2010}
K.~Guo, P.~Ishwar, and J.~Konrad.
\newblock Action recognition using sparse representation on covariance
  manifolds of optical flow.
\newblock In {\em International Conference on Advanced Video and Signal Based
  Surveillance (AVSS)}, pages 188--195, 2010.

\bibitem{KaiGuo2013}
K.~Guo, P.~Ishwar, and J.~Konrad.
\newblock Action recognition from video using feature covariance matrices.
\newblock {\em IEEE Transactions on Image Processing}, 22(6):2479--2494, 2013.

\bibitem{Harris88}
C.~Harris and M.~Stephens.
\newblock A combined corner and edge detector.
\newblock In {\em Alvey Vision Conference}, pages 147--151, 1988.

\bibitem{Hoai2011}
M.~Hoai, Z.-Z. Lan, and F.~De~la Torre.
\newblock Joint segmentation and classification of human actions in video.
\newblock In {\em Conference on Computer Vision and Pattern Recognition
  (CVPR)}, pages 3265--3272, 2011.

\bibitem{Laptev2005}
I.~Laptev.
\newblock On space-time interest points.
\newblock {\em International Journal of Computer Vision}, 64(2-3):107--123,
  Sept. 2005.

\bibitem{ChuanZhen2014}
C.~Li, B.~Su, J.~Wang, H.~Wang, and Q.~Zhang.
\newblock Human action recognition using multi-velocity {STIP}s and motion
  energy orientation histogram.
\newblock {\em Journal of Information Science and Engineering}, 30:295, 2014.

\bibitem{Mendoza2007}
M.~A. Mendoza and N.~P\'{e}rez de~la Blanca.
\newblock {HMM}-based action recognition using contour histograms.
\newblock In {\em Pattern Recognition and Image Analysis}, volume 4477 of {\em
  Lecture Notes in Computer Science}, pages 394--401. Springer Berlin
  Heidelberg, 2007.

\bibitem{KReddy2013}
K.~K. Reddy and M.~Shah.
\newblock Recognizing 50 human action categories of web videos.
\newblock {\em Machine Vision and Applications Journal (MVAP)}, pages 971--981,
  2013.

\bibitem{Reddy_TCSVT_2013}
V.~Reddy, C.~Sanderson, and B.~Lovell.
\newblock Improved foreground detection via block-based classifier cascade with
  probabilistic decision integration.
\newblock {\em IEEE Transactions on Circuits and Systems for Video Technology},
  23(1):83--93, 2013.

\bibitem{Sanderson_NICTA_2010}
C.~Sanderson.
\newblock Armadillo: An open source {C}++ linear algebra library for fast
  prototyping and computationally intensive experiments.
\newblock Technical report, NICTA, 2010.

\bibitem{AndresSanin2013}
A.~Sanin, C.~Sanderson, M.~Harandi, and B.~Lovell.
\newblock Spatio-temporal covariance descriptors for action and gesture
  recognition.
\newblock In {\em IEEE Workshop on Applications of Computer Vision (WACV)},
  pages 103--110, 2013.

\bibitem{Schuldt2004}
C.~Schuldt, I.~Laptev, and B.~Caputo.
\newblock Recognizing human actions: {A} local {SVM} approach.
\newblock In {\em International Conference on Pattern Recognition (ICPR)},
  volume~3, pages 32--36, 2004.

\bibitem{Qinfeng2008}
Q.~Shi, L.~Wang, L.~Cheng, and A.~Smola.
\newblock Discriminative human action segmentation and recognition using
  semi-{M}arkov model.
\newblock In {\em Conference on Computer Vision and Pattern Recognition
  (CVPR)}, pages 1--8, 2008.

\bibitem{Shimosaka2007}
M.~Shimosaka, T.~Mori, and T.~Sato.
\newblock Robust action recognition and segmentation with multi-task
  conditional random fields.
\newblock In {\em International Conference on Robotics and Automation}, pages
  3780--3786, 2007.

\bibitem{Wagner1995}
D.~B. Wagner.
\newblock Dynamic programming.
\newblock {\em The Mathematica Journal}, pages 42--51, 1995.

\bibitem{HengWang2013Journal}
H.~Wang, A.~Kl\"{a}ser, C.~Schmid, and C.-L. Liu.
\newblock Dense trajectories and motion boundary descriptors for action
  recognition.
\newblock {\em International Journal of Computer Vision}, 103(1):60--79, 2013.

\bibitem{HengWang2013iccv}
H.~Wang and C.~Schmid.
\newblock Action recognition with improved trajectories.
\newblock In {\em International Conference on Computer Vision (ICCV)}, pages
  3551--3558, 2013.

\bibitem{Borzeshi2012}
E.~Zare-Borzeshi, O.~Perez-Concha, and M.~Piccardi.
\newblock Human action recognition in video by fusion of structural and
  spatio-temporal features.
\newblock In {\em Structural, Syntactic, and Statistical Pattern Recognition},
  volume 7626 of {\em Lecture Notes in Computer Science}, pages 474--482.
  Springer Berlin Heidelberg, 2012.

\end{thebibliography}
\end{document}